\documentclass[sigconf]{acmart}
\acmYear{2022} 
\usepackage{hhline}
\usepackage{stfloats}
\usepackage{algorithm}
\usepackage{algorithmic}
\usepackage{setspace}
\usepackage{amsmath}
\usepackage{autobreak}
\usepackage{mathrsfs}
\usepackage{bm}
\usepackage{graphicx}
\usepackage{float}
\usepackage{subfigure}
\usepackage{caption}
\usepackage{multicol}
\usepackage{multirow}
\usepackage{booktabs}
\usepackage{xcolor}
\usepackage{makecell}

\usepackage{balance}
\begin{document}
% make the title area

\title{An Embedded Feature Selection Framework for Control}

\author{Jiawen Wei}
\authornote{Both authors contributed equally to this research.}
\affiliation{%
  \institution{Central South University}
  \city{Changsha}
  \state{Hunan}
  \country{China}}
\email{jiawennw@gmail.com}

\author{Fangyuan Wang}
\authornotemark[1]
\affiliation{%
  \institution{Zhejiang Sci-Tech University}
  \city{Hangzhou}
  \state{Zhejiang}
  \country{China}}
\email{wfy11235813@gmail.com}

\author{Wanxin Zeng}
\affiliation{%
  \institution{Central South University}
  \city{Changsha}
  \state{Hunan}
  \country{China}}
\email{194611090@csu.edu.cn}

\author{Wenwei Lin}
\affiliation{%
  \institution{Central South University}
  \city{Changsha}
  \state{Hunan}
  \country{China}}
\email{linwweix@gmail.com}

\author{Ning Gui}
\authornote{Corresponding Author.}
\affiliation{%
  \institution{Central South University}
  \city{Changsha}
  \state{Hunan}
  \country{China}}
\email{ninggui@gmail.com}

\renewcommand{\shortauthors}{Jiawen ad Fangyuan, et al.}

\begin{abstract}

Reducing sensor requirements while keeping optimal control performance is crucial to many industrial control applications to achieve robust, low-cost, and computation-efficient controllers. However, existing feature selection solutions for the typical machine learning domain can hardly be applied in the domain of control with changing dynamics. In this paper, a novel framework, namely the Dual-world embedded Attentive Feature Selection (D-AFS), can efficiently select the most relevant sensors for the system under dynamic control. Rather than the one world used in most Deep Reinforcement Learning (DRL) algorithms, D-AFS has both the real world and its virtual peer with twisted features. By analyzing the DRL's response in two worlds, D-AFS can quantitatively identify respective features' importance towards control. A well-known active flow control problem, cylinder drag reduction, is used for evaluation. Results show that D-AFS successfully finds an optimized five-probes layout with 18.7\% drag reduction than the state-of-the-art solution with 151 probes and 49.2\% reduction than five-probes layout by human experts. We also apply this solution to four OpenAI classical control cases. In all cases, D-AFS achieves the same or better sensor configurations than originally provided solutions. Results highlight, we argued, a new way to achieve efficient and optimal sensor designs for experimental or industrial systems. Our source codes are made publicly available at https://github.com/G-AILab/DAFSFluid.
\end{abstract}




\maketitle

\section{Introduction}

For experimental and industrial systems, reducing sensor requirements while maintaining optimal control performance is crucial for human experts to understand target systems and potentially transpose these techniques to industrial cases. It is one of human beings' most critical intellectual activities, i.e., extracting knowledge from unknown environments to recognize and exploit the world\cite{sutton2018reinforcement}. To better understand the target system, scientists/engineers use intuition and domain knowledge to identify critical information based on continuous observations of system performance\cite{khurana2018feature}. 
But for a complex system with massive sensory inputs, this handcrafted solution through trial-and-error is normally time-consuming, costly, and heavily reliant on human expertise.
Thus, it is vital to develop an effective and systematic solution to identify relevant sensors without or with minimal human intervention.

Nowadays, DRL has shown great success in complex system control problems\cite{lillicrap2016continuous} by combining reinforcement learning and deep learning\cite{mnih2015human}. The introduction of Deep Neural Networks (DNN) can extract important features from a huge set of raw inputs. Despite its success, the opaque nature of DNN hides the critical information about which sensors are most relevant for control\cite{samek2017explainable,du2018towards}. Existing solutions for sensor selection under DRL have been limited to the recursive sensor set selection and evaluation with certain heuristic algorithms\cite{paris2021robust}. 
As their complexity increases rapidly with the number of candidate sensors, these methods are inapplicable for systems with massive possible inputs.

\begin{figure}
    \centering
    \includegraphics[width= 0.48\textwidth]{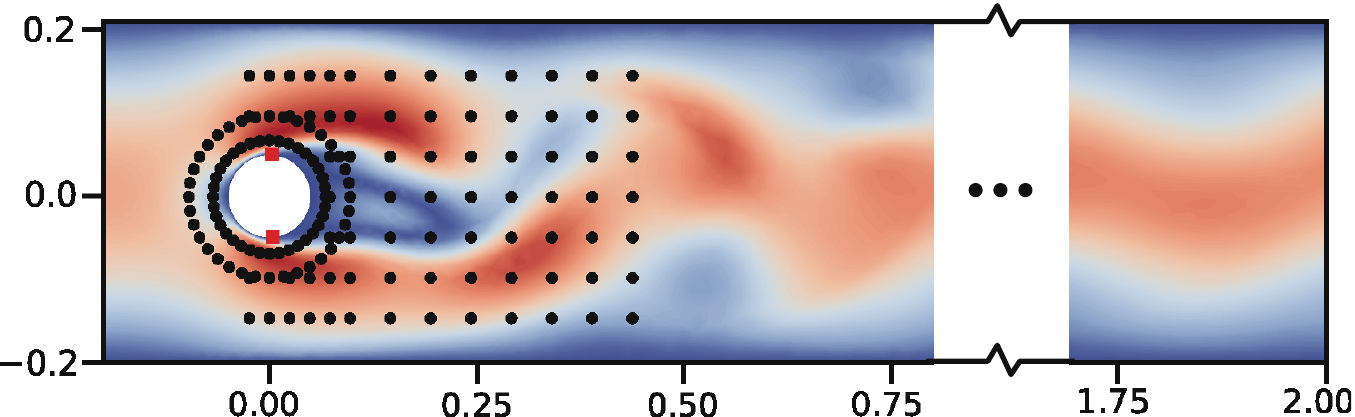}
    \caption{Instantaneous velocity flow field with 151 probes. Black dots represent sensor locations and colors represent velocity(fields).}
    \label{fig:intro}
    \vspace{-0.2in}
\end{figure}

This requirement is similar to the Feature Selection (FS) problem in Machine Learning(ML) \cite{li2017feature}, also known as variable selection or attribute selection. This process aims to select a subset of relevant features with respect to the modeling labels. 
% A subset of relevant features is selected for model construction according to the performance of a learning algorithm or model, and labels are demanded to guide the feature selection process.
However, due to the distinctiveness of control process, choosing a suitable variable as a label is very troublesome. 
For example, Fig.~\ref{fig:intro} shows a classic flow control problem to reduce the drag and lift of a cylinder by adjusting flow injection/suction with two jets(represented by two red dots) and 151 probes. 
Firstly, the immediate reward corresponding to the time series state set is often meaningless as a label. The instance injection/suction might not change the immediate reward significantly. Secondly, the goal for control typically uses the discounted cumulative return as a loss function that often has long and uncertain latency and has little relevance to the single-step state. 
Thus, neither immediate reward nor cumulative return is appropriate as the mandatory training label in most feature selection methods. In later experiments, we also experimentally prove that existing feature selection methods are not suitable in systems with control.

For effective control, a DRL controller has already learned the relevance of different inputs during its exploration and exploitation. Can we make the hidden knowledge on the relevance of respective sensors explicit and provide qualitative evaluation during DRL training? To address this challenge, we propose a new solution, the Dual-world based Attentive embedded Feature Selection(D-AFS), a general mechanism that enables the DRL controller to interact with the real world and its virtual peer with twisted features. 
Our method combines general DRL framework and feature selection module, which can identify the contribution of inputs simultaneously with strategy learning process. In summary, our contributions are:

%  Currently, those tasks are limited mainly to human expertise and depend on hand-crafted experiments. 
\begin{itemize}
    \item \textbf{A new direction for embedded feature selection in control}: We combine the exploration and exploitation capabilities of DRL with an attention-based feature evaluation module, that qualitatively evaluates the impact of different features by translating feature weight calculation to feature selection probability problem. The method can identify the feature importance of unknown systems for control during DRL's strategy learning process.
    \item \textbf{A dual-world architecture design}: We propose a dual-world architecture that includes real world and virtual world, of which the virtual world is mapped from its real peer by the feature evaluation module. Dual-world enables D-AFS complete feature importance ranking without jeopardizing DRL's strategy exploration. 
    % A dual-world architecture with the traditional real world and the virtual world twisted from its real peers is proposed to complete the feature selection task without jeopardizing the DRL control task. An attention-based feature evaluation mechanism is proposed to qualitatively evaluate the impact of different features by translating the feature weight calculation to the feature selection probability problem.
    \item \textbf{Excellent results on active flow control}: We apply D-AFS with the Proximal Policy Optimization (PPO) algorithm\cite{schulman2017proximal} in the case mentioned above. D-AFS/PPO successfully reduces a five-sensor layout from 151 possible probes, with a drag coefficient of 2.938, very close to the theoretical limit of 2.93 without vortex shedding, and achieves 18.7\% drag reduction than the state-of-the-art solutions using 151 probes.
\end{itemize}

We also successfully adapt Deep Q-Network(DQN)\cite{mnih2015human} and Deep Deterministic Policy Gradient (DDPG)\cite{lillicrap2016continuous} into D-AFS, apply them on four OpenAI Gym\cite{brockman2016openai} classical control cases. In all cases, D-AFS selects the same or better sensor subsets than the original expert-based solutions (Gym provided). Results clearly show that our solution can effectively identify system key sensors, improve the interpretability of the target system and reduce the reliance on human expertise. 

% The detailed simulation results and source codes are attached in the supplementary materials.  

% Experiment results show that our proposed method achieves . Compared to the five-sensor layout by human experts, D-AFS achieves 252.9\% increase in the recirculation area and 37.5\% drag reduction. 

% The rest of this paper is organized as follows: Section II discusses related work. Then the proposed dual system and the major modules in D-AFS are introduced in Section III. Section IV integrates an attention-based module into two existing DRL algorithms to form the concrete algorithms of D-AFS. Then experiments are performed, and the results are analyzed in Section V. The final section concludes our work and points out the possible directions for future work.

\section{Related Work}
This section discusses the related work from two aspects: 1) The feature selection methods in machine learning; 2) The abbreviated research for sensor selection in active flow control.

\subsection{Feature Selection in ML}
Feature selection is a technology that contains automatically discovering, evaluating, and selecting relevant features\cite{bengio2013representation}. 
Generally, it can be divided into three categories\cite{li2017feature}:
1) Filter methods separate feature selection from the learning process and only rely on the measures of the general characteristics of the training data to evaluate the feature weights. Different feature selection algorithms exploit various types of criteria to define the relevance of features, e.g., Fisher score\cite{gu2011generalized} and ReliefF\cite{robnik2003theoretical}. 2) Wrapper methods\cite{tang2014feature}  predict accuracy of a predefined learning algorithm to evaluate the quality of recursive selected features. They normally suffer high computation costs and low scalability. 3) Embedded methods depend on the interactions with the learning algorithm and evaluate feature sets according to the interactions. The most recent approaches are in this class e.g., XGBoost\shortcite{chen2016xgboost}, LightGBM\shortcite{ke2017lightgbm} and Random Forest(RF)\shortcite{breiman2001random}. Recently, DNN-based solutions are proposed,e.g., the attention-based, AFS\shortcite{gui2019afs} or casualty-based \cite{Yu2021Unified}. However, trajectories generated by the DRL controller do not follow the independent and identically distributed assumption, making it impossible to directly transfer and use the feature selection methods in the ML domain.
% \textcolor{red}{the uniqueness of the DRL mechanism makes it impossible to directly transfer and use the feature selection methods in the ML domain.} \textcolor{blue}{There's independent and identically distributed hypothesis in machine learning.}
\cite{hu2019design} creates virtual sensors to do fault detection by using existing sensors and applying domain knowledge to the data. However, this approach does not actually reduce the number of sensors and cannot be applied to control tasks.
As pointed out in the introduction, for model-free systems, it is challenging to provide appropriate labels for each system state, as demanded in most ML feature selection approaches. Thus, it is difficult for existing ML feature selection methods to be applied in the DRL control environment.

% \noindent\textbf{Policy Gradient Method}
% \vspace{-0.2in}
% Policy gradient methods update the policy $\pi$ to the direction of the expected total reward gradient $g = \nabla_{\theta}\mathbb{E}_{\pi}[R_{\tau}]$, 
% where $\tau$ is a trajectory generated by sampling actions according to the policy $\pi$\cite{sutton2018reinforcement}.
% It is also reasonable to learn a value function estimation in addition to the policy $\pi$. And that is exactly what the Actor-Critic does. 
% PPO\cite{schulman2017proximal} is popular Actor-Critic methods among them. It proposes a novel method that forms a lower bound estimate of the policy performance and use a clipped surrogate objective to constrain the size of policy update.
% This guarantee that the policy will not update too large to stuck in the local optimal while still proceeding towards the direction of gradient descent.
\subsection{Optimal Sensor Placement in Flow Control}
% \vspace{-0.1in}
The issue of optimal sensor placement has been investigated with certain domain-specific knowledge. Most existing solutions,e.g., \cite{bright2013compressive,cohen2006heuristic} are based on compressed sensing theory that selects a limited number of sensors that can effectively perform flow reconstruction.  These approaches do not depend on the learning algorithm and can be seen as a special filter feature selection algorithm. This stream of methods also has very high computation complexity, as we need to reconstruct the whole flow. Furthermore, as stressed by Stephan \& Simon \shortcite{oehler2018sensor}, control does not systematically require faithful flow reconstruction. The partial knowledge of relevant `hidden' variables may be sufficient. 

Despite its success in many domains, only in the recent work of Rabault et al. \shortcite{rabault2019artificial},  DRL has been used to control the flow past a 2D cylinder. They use 151 probes to learn control strategy by PPO in the 2D Kármán Vortex Street simulation. Tang et al. \shortcite{tang2020robust} further improve their work through extending the Reynolds numbers range and sensor observations. These researches have not yet involved the optimization of sensor selection. Paris et al. \shortcite{paris2021robust} recently proposes a sensor selection algorithm based on a well-trained PPO controller. However, this approach adopts the so-called wrapper method by evaluating different combinations of features. Their work has very high computation complexity, and their discussion is limited to a system with 15 probes.

To the best of our knowledge, there have been no systematic solutions yet to obtain interpretable environmental representations through the learning and exploration processes of DRL.

\section{Definitions and Notations}
This section gives a general definition of the problem studied in this paper and briefly demonstrates the related notations. 

\noindent\textbf{Raw Features.} Given a controlled object, the basic observations are a set of states periodically collected by sensors deployed in the system. We use $\bm{\mathcal{S}}_r =\{s^k|k=1,2,...,m\}$ to describe the set of raw states, where $k$ represents the number of sensors. 

\noindent\textbf{Feature Selection for Control.}
The goal of feature selection for control under the DRL framework is stated as follows. Given a set of features $\bm{\mathcal{S}}_r$ from the real world, find a subset of features, $\bm{\mathcal{S}}^* \subseteq \bm{\mathcal{S}}_r$, to maximize the expected cumulative return $\mathcal{R}_t$ for a given DRL algorithm.
As shown in Eq. (\ref{eq:optimal_s}):
\begin{equation}
% \vspace{-0.1in}
\label{eq:optimal_s}
    \bm{\mathcal{S}}^* = \arg\max_{\bm{\mathcal{S}}_r} \mathcal{R}_t = \arg\max_{\bm{\mathcal{S}}_r} \sum_{t}^{\infty}\gamma^{t}r_{t}
\end{equation}
where $\gamma$ is a discount rate.

\noindent\textbf{Discussion.} For a target system, if a well-designed controller is capable of achieving effective control, the raw features must contain all the necessary information for control.
% it is logical to assume that raw features contain all the necessary information for control if a well-designed controller can achieve effective control.
However, it has been wildly agreed that reducing sensor requirements while keeping optimal control performance is crucial to many industrial control applications to achieve robust, low cost, and computation efficient controllers\cite{khurana2018feature}. 
Previous works on the industrial control with DRL\cite{paris2021robust} demonstrate that when the selected feature representation matches the form required by the physical model, the DRL controller can usually even achieve a better control effect. 
This characteristic is of vital importance in understanding the underlying physical model of the target systems.

\begin{figure*}[tb]
    \centering
    \includegraphics[width=0.85\textwidth]{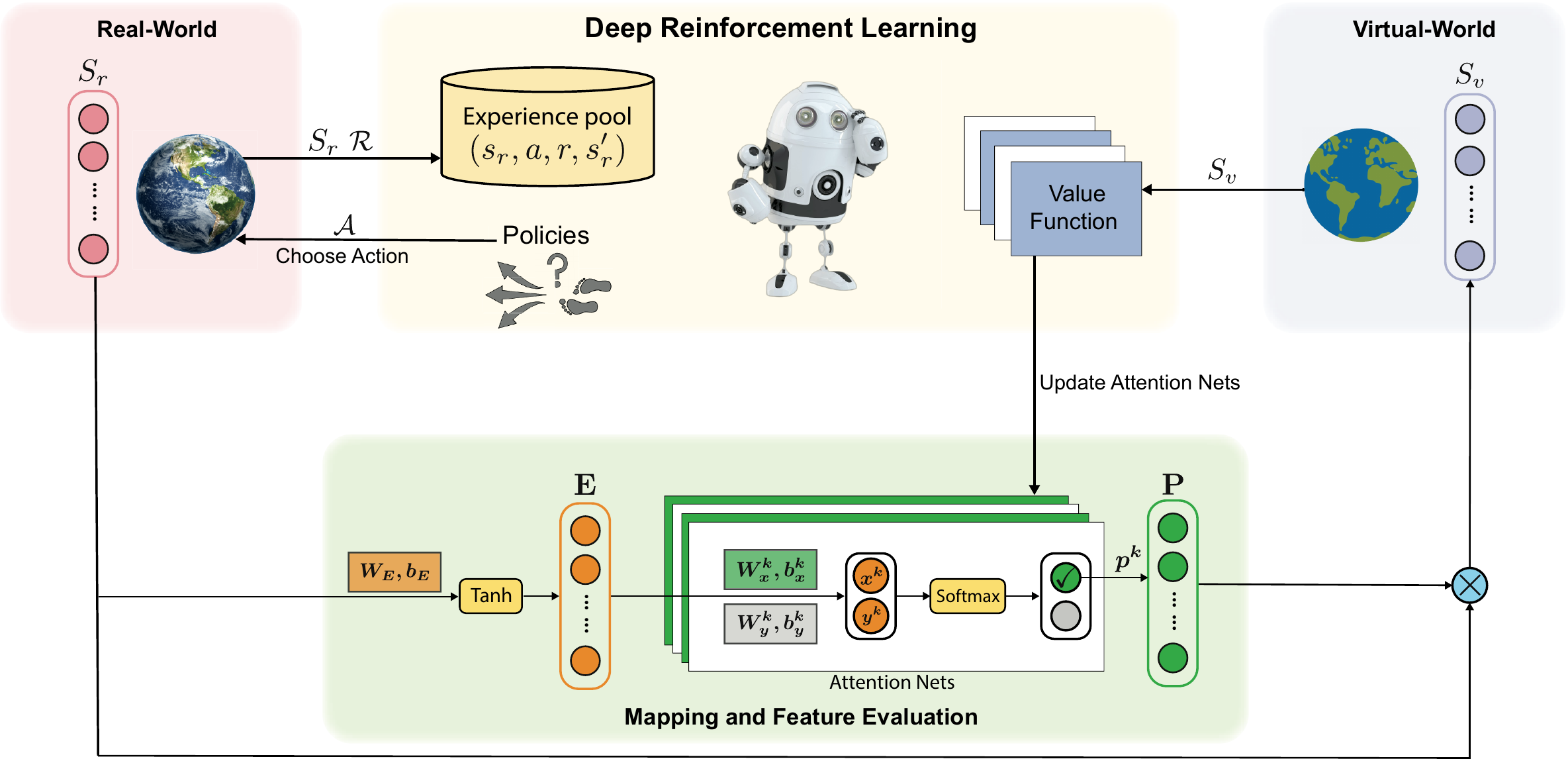}
    \caption{The overall architecture of D-AFS with the ``real'' and ``virtual'' world environment, the evaluation module and the DRL module. The virtual world is mapped from the real world. DRL learns in two mixed worlds, explores the correct control strategy in the real world, and provides evaluation information for feature selection in the virtual world.}
% \vspace{-0.15in}
    \label{fig:architecture}
\end{figure*}

\section{Proposed Method}

In this section, the overall structure of D-AFS and the mechanism of embedded feature selection are illustrated and analyzed.

\subsection{Design Principles}

The basic design principle for two different worlds is rather straightforward. Firstly, if a feature is irrelevant or weakly related to the system, it has little effect on a ``good'' control strategy. Conversely, a feature strongly related to the physical model of the system or the control goal should have a pronounced impact on the control strategy.  Generally, the one that is more benign for the controller might have more impacts. 

However, the dual-world design brought a series of problems to the existing DRL solutions with only one world. Three main issues need to be tackled: 1) How to effectively combine the real and virtual world without affecting the normal operation of DRL? 2) How to dynamically adjust the combination of features in the virtual world space without jeopardizing the DRL learning process? 3) How to modify existing DRL algorithms to allows online feature evaluation?

Thus, we introduce the dual-world architecture, which contains three main modules, the environment module, the evaluation module, and the DRL module, as shown in Fig. \ref{fig:architecture}. The environment and the DRL module are inherited directly from the original DRL framework, but detailed designs are slightly different to support the two worlds. The evaluation module calculates the importance of each feature through the attention mechanism and then helps generate the virtual world.
The DRL module, evaluation module, and the unique dual-world system are loosely coupled in the architecture we designed. 
We minimize the modification of the general structure of traditional DRL algorithms and maintain its integrity to the greatest extent, which significantly promotes the reuse of the existing state-of-the-art DRL algorithms. Therefore, the DRL module can be determined according to the specific problem.

% \noindent\textbf{Dual-world, DRL and Evaluation Module }

% To address these three questions, we first need to clarify the different roles between the real world and the virtual world of the dual-world space. On the one hand, we need the real world to provide system information to explore the correct control strategy. On the other hand, the virtual world searches for the optimal features and representations by continuously tuning the mapping from the real world to the virtual world. 

% The DRL module, evaluation module, and the unique dual-world system are loosely coupled in the architecture we designed. 
% We minimize the modification of the general structure of the traditional DRL algorithm and maintain its integrity to the greatest extent, which significantly promotes the reuse of the existing state-of-art DRL algorithm. Therefore, DRL modules can be determined according to the specific problem. 

% \textcolor{red}{
% The dual-world environment is proposed to tackle the first issue and is illustrated in Sec.~\ref{sec:dual-world}. 
% The evaluation and virtual world generation are performed in the Evaluation module (described in Sec.~\ref{sec:AE}) to address the second issue. 
% The final issue is alleviated by design proposed in Sec.~\ref{sec:ppo}. 
% }

% \vspace{-0.2cm}
\subsection{Dual-world}
\label{sec:dual-world}
We first need to clarify the different roles between the real world and the virtual world of the dual-world space. On the one hand, we need the real world to provide system information to explore the correct control strategy. On the other hand, the virtual world searches for the optimal features and representations by continuously tuning the mapping from the real world to the virtual world. 
The real world features are the system's original features $\bm{\mathcal{S}}_r$, which is used to interact with the policy of the DRL module.
And the virtual world is the real world mapping through the evaluation module, which can be used to evaluate the validity of the selected features. The mapping function is generated with the attention-based evaluation module illustrated in the following section.  In D-AFS, the environment is the combination of the real and virtual world $\bm{\mathcal{S}}_r \cup \bm{\mathcal{S}}_v $.

\subsection{Attention-based Evaluation Module}
\label{sec:AE}

The attention-based evaluation (AE for short) module aims to evaluate the importance of different features by assigning different weights (denoted as $\bm{P}$). The structure of AE is shown in the green area in Fig. \ref{fig:architecture}.

The AE module is a revised attention mechanism extended from our previous work\cite{gui2019afs}. This module consists of two key components, a dense network $\bm{E}$ with $N_E$ neuronal units and several Attention Networks (AN for short) for calculating the importance weights of each input.
% $\bm{E}=\tanh{({\bm{\mathcal{S}}}_r}^{\rm T}\bm{W}_E+\bm{b}_E)}$

The dense network \(\bm{E} = \tanh({\bm{\mathcal{S}}_r}^{\rm T} \bm{W}_E + \bm{b}_E)\), \(\bm{W}_E\in\mathbb {R}^{m\times N_E}\), \(\bm{b}_E\in\mathbb{R}^{N_E}\), is used to abstract the internal relationship between the inputs $\bm{\mathcal{S}}_r$. $\bm{W}_E$ and $\bm{b}_E$ are parameters, nonlinear function $\tanh(\cdot)$ keeps positive and negative values. To some extend, the dense network can reduce the redundancy of all sensors, of which the units $N_E$ can be adjusted according to respective inputs.

Once we have extracted the relationship of features by $\bm{E}$, we configure an attention network for each feature to generate weight separately. We do not adopt the typical soft attention mechanism to generate a weighted arithmetic mean of all features. This operation might result in small weights for most features and large weights for a tiny number of features and suffer the loss of details on the whole feature sets.
These weights can determine whether one feature should be selected or not. More detailed, it generates two values, $\bm{X}=\bm{E}\bm{W}_x+\bm{b}_x$ and $\bm{Y}=\bm{E}\bm{ W}_y+\bm{b}_y$, to represent the selected/unselected probability of a feature, respectively. 
In general, the higher the weight of the feature being selected ($\bm{X}$), the more important the feature is for control. 
Then we use the softmax function to magnify the distance between selected and unselected.
\begin{equation}
    \label{eq:softmax}
    p^k=\frac{\exp(x^k)}{\exp{\left(x^k\right)}+\exp(y^k)},x^k\in\bm{X},y^k\in\bm{Y}
\end{equation}
Finally, these attention nets will generate a matrix of attention weights $\bm{P} =\left \{p^k|k=1,2,...,m\right\}$, which has the same dimension as features. 
With Eq. \ref{eq:att}, we generate a feature representation of the virtual world with $\bm{\mathcal{S}}_r$ multiplied by attention weights.
\begin{equation}
    \bm{\mathcal{S}}_v=\bm{\mathcal{S}}_r \odot \bm{P}
\label{eq:att}
\end{equation}
The AE module generates a dynamic mapping function that maps representations from the real world $\bm{\mathcal{S}}_r$ to the virtual world $\bm{\mathcal{S}}_v$, which will be used for DRL to update the strategy. Additionally, with the attention weights matrix $\bm{P}$ generated by AE, we can evaluate respective features when DRL has explored an efficient strategy.
% {\color{red}
% The virtual world $\bm{\mathcal{S}}_v$ will be used for the DRL model to update strategy and the AE module actually generate a dynamic mapping function that maps the real world representation to the virtual world. How to update the parameters of the AE module will be explained later. When the system reaches a good control state, the importance of respective features can be analyzed according to $\bm{P}$ generated by AE.
% }

% \vspace{-0.2cm}
\subsection{PPO in Dual-world}
\label{sec:ppo}
As far as we know, value functions are used in one way or another among the current model-free DRL algorithms\cite{SpinningUp2018}.
In our algorithm, in addition to the role of the original DRL algorithm, the value function also needs to guide the AE module to select valuable features during the training process. This subsection shows how to integrate the D-AFS with the popular Proximal Policy Optimization(PPO) algorithm, denoted as D-AFS/PPO. PPO is a popular policy-based online algorithm with clipped probability ratios that forms a lower bound of the performance of the policy.
In order to allow the agent to interact with the environment correctly, we use the real world as the input of the PPO actor network, which is the same as the original PPO algorithm.
Thus, the actor network with parameters $\theta$ can be updated by:

\begin{equation}
    \begin{aligned}
        \theta^* = \arg \max_{\theta} L^{CLIP}(\theta)
    \end{aligned}
    % L_t(\theta)=min(r_t(\theta_k) A^{\pi_{\theta_k}}(s_t, a_t), clip(r_t(\theta)), 1-\epsilon, 1+\epsilon)A^{\pi_{\theta_k}}(s_t, a_t)
    % \theta_{k+1} = \arg \max_{\theta} \frac{1}{|\mathcal{D}_k|T} \sum_{\tau \in \mathcal{D}_k} \sum_{t=0}^T min(r_t(\theta))A^{\pi_{\theta_k}}(s_r,a)
    \label{eq:actor}
\end{equation}
where,
\begin{equation*}
    L^{CLIP}(\theta)=\mathbb{E}_t[\min(r_t(\theta), {\rm clip}(r_t(\theta), 1-\epsilon, 1+\epsilon))A^{\pi_{\theta_t}}(s_r,a)],
\end{equation*}
$r_t(\theta)$ is the probability ratio of the old and the new policy, the subscript $t$ is the number of iteration.
% and $r_t(\theta)=\frac{\pi_{\theta}(a_t | s_{r}_{t})}{\pi_{\theta}_{old}(a_t | s_{r}_{t})}$.

The difference is that in our algorithm, we use the virtual world as the input of the critic network, allowing the critic network to evaluate the goodness of the selected features of the evaluation module while scoring the performance of the actor network. That is:
\begin{equation}
    \phi^*, \psi^*=\arg \min_{\phi,\psi} \mathbb{E}_t (V_{\phi,\psi}(s_v)-\mathcal{R}_t)^2
    \label{eq:critic}
\end{equation}
$\phi$ and $\psi$ are parameters of critic network and attention module, respectively. $V_{\phi, \psi}(s_v)$ is value function and $\mathcal{R}_t$ is accumulative return.

The detailed structure of this D-AFS/PPO is shown in Algorithm \ref{algorithm}. The bold pseudo-codes are the revised parts to support the dual-world DRL learning. 
\vspace{-0.1in}
\begin{algorithm}
    \caption{D-AFS/PPO, Actor-Critic based}
    \label{algorithm}
    \begin{algorithmic}
        \STATE Initial parameters of actor network $\theta_0$, critic network $\phi_0$ and attention network $\psi_0$.
        \FOR{$t=1,2,...$}
        \STATE Run policy $\pi(\theta_{t})$ to get trajectories $\mathcal{D}_t(s_r, a, r, s^\prime_r)$
        \STATE Compute accumulative return $\mathcal{R}_t$
        \STATE \textbf{Compute attention weights $\bm{p}$ based on Eq. (\ref{eq:softmax})}
        \STATE Compute $\bm{s_v \gets s_r \odot p}$
        \STATE Compute advantage estimates $A_t$ based on current function $\bm{V_{\phi_t, \psi_t}(s_v)}$
        \STATE Update the actor network by maximizing the PPO-Clip based on Eq. (\ref{eq:actor}) with Adam $\theta_{old} \gets \theta$
        \STATE Update the critic network by regression on MSE based on Eq. (\ref{eq:critic}) with Adam $\{ \phi_{old} \gets \phi, \bm{\psi_{old} \gets \psi} \}$
        \ENDFOR
    \end{algorithmic}
\end{algorithm}

% Since the virtual world as the input of the critic network in our architecture, the parameter of attention networks $\theta^A$ is updated while optimizing the critic network by minimizing the loss function as follows:

% $r_t$ is the reward returned by the current step, and $\gamma$ is a discounting factor, $y_i$ is the output of the target network of the critic.

% It can be known from the Fig. \ref{fig:ddpg} that the input of all policy functions in the actor network is the real world environment, so when the network parameters of the actor network are updated, the parameters of attention nets ${\theta}^A$ remain unchanged. Besides the parameters updating mentioned above, the target network parameters' updating methods are the same as those in the conventional DDPG, using the soft update method.

% It is worth noting that our modifications are mainly about the computation of the value function, which uses the virtual world as input. These modifications are designed as non-intrusive as possible so that it can be easily extended to various DRL algorithms. Therefore, our structure can make full use of the advancement of the DRL algorithm and identify the key features of the system at the same time. 
% \vspace{-0.1in}
\section{Experiments and Analysis}

In this section, the experimental system and settings are firstly introduced. Then the results for both FS solutions and DRL solutions are provided and analyzed.
We also briefly provide further experiments in OpenAI Gym to demonstrate D-AFS's adaptability. 

\subsection{System Description and Settings}

 Drag reduction and active flow control are critical technologies in the fluid industry. Here, we briefly introduce the target system and experiment settings. 
 
 \noindent\textbf{Target System: } This environment is from a well-known benchmark \cite{schafer1996benchmark} which consists of a cylinder of diameter $D=0.1$ immersed in a constant flow of a box of total length $L=2.2$ and height $H=0.41$. The cylinder is slightly downward from the center of the Y-Axis by 0.005. The mean velocity magnitude is $\Bar{U}=1$.
And the Reynolds number based on the mean velocity magnitude and cylinder diameter ($Re=\Bar{U}D/v$, with $v$ the kinematic viscosity) is set to $Re=100$.
When the target system is in an unstable state, the flow velocity and pressure of the wake behind the cylinder show periodic changes due to the transmission of the vortex street. As the learning of the control strategy becomes effective, the velocity and the pressure of the wake gradually decrease.

\begin{figure}[th]
    \centering
    \includegraphics[width=0.48\textwidth]{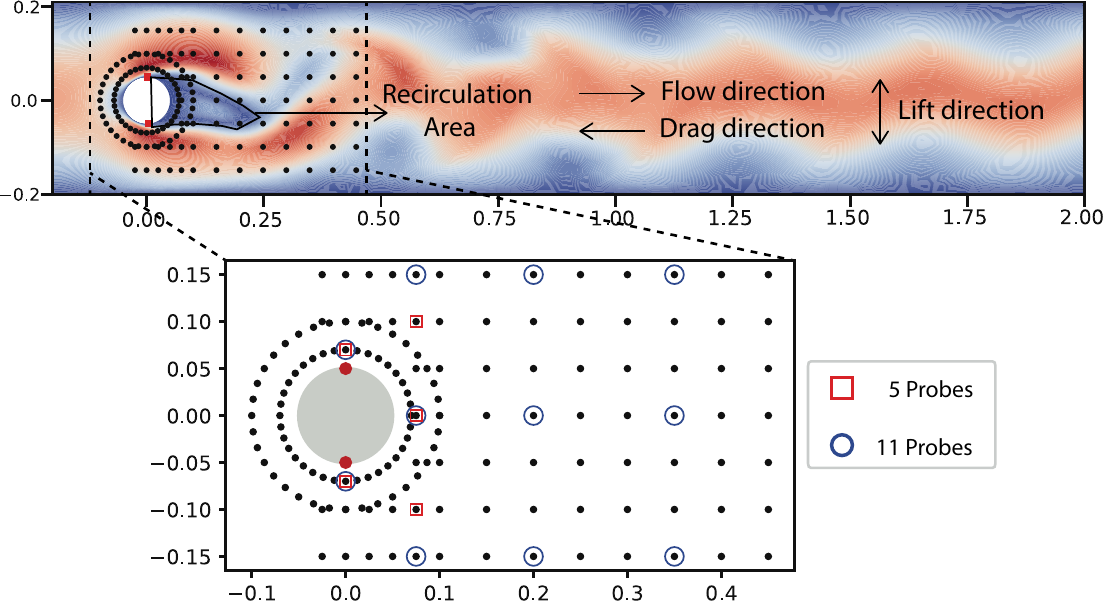}
    \caption{The 2D active flow control simulation environment without control. The location of the jets is indicated by the red dots. }
    \label{fig:cfd_env}
\end{figure}

Two synthetic jets are located at the top/bottom extremities of the cylinder, which we can use DRL to control, as indicated in Fig. \ref{fig:cfd_env} with solid red dots. There are 151 velocity probes placed around the cylinder and the near weak, which provide the network with a relatively detailed flow description, as indicated in Fig.~\ref{fig:cfd_env} with black dots. Circle and Square are Top5 and Top11 probes selected by human experts. Our experimental environment is inherited from \cite{rabault2019artificial}. Detailed settings and calculations can refer to the paper.\footnote{https://github.com/jerabaul29/Cylinder2DFlowControlDRL}.

% \begin{figure}[ht]
%     \centering
%     \includegraphics[width=0.4\textwidth]{pic/paperprobes.eps}
%     \caption{The location of all 151 probes.  The 5/11 probes selected by \cite{rabault2019artificial}.}
%     \label{fig:my_label}
% \end{figure}

% Figure \ref{fig:cfd_p_v}, top a. and b. show the unsteady pressure and velocity fluctuations induced by the vortexes without active control.

\noindent\textbf{D-AFS/PPO Settings: } Due to the limited continuous action space of the 2D flow control system, we use Beta policy\cite{Chou-2017-26161}, which has finite support and no probability density that falls outside the boundary, to sample actions in the training process and use deterministic policy in the testing process. It is also the default policy of Tensorforce\cite{tensorforce} for limited continuous action space, which is used by \cite{rabault2019accelerating}.
The critic network consists of 4 layers: an input layer, two hidden layers with 512 cells, and an output layer.
In the AE module, the dense network \(N_E\) has 20 cells.

The original paper\cite{rabault2019artificial} takes 24 hours to train a stable controller on a single CPU.
And about 99.7\% of the computation time is spent on the environment itself using FEniCS\cite{alnaes2015fenics} rather than the DRL algorithm \cite{rabault2019accelerating}.
Thus, we designed the asynchronous training architecture to accelerate Algorithm \ref{algorithm} by syncing parameters and averaging gradients on each worker. The architecture reduced the training time from 40 hours to 3.6 hours by using 20 threads.

\noindent\textbf{Evaluation Metrics and Reward Function:} We use drag coefficient $C_D$, lift coefficient $C_L$, recirculation area size, and velocity fluctuations and pressure fluctuations to evaluate the performance of a trained controller. The drag coefficient and lift coefficient are the normalized total instantaneous drag (opposite the constant flow direction) and lift (perpendicular to the constant flow direction) on the cylinder. And the recirculation area is the region in the downstream neighborhood of the cylinder where the horizontal component of the velocity is negative. To compare fairly, we use the standard PPO with the absolute same configuration for learning as the arbiter to evaluate different probe configurations. 

In short, an excellent active flow control strategy means multiple evaluations, small $C_D$ and $C_L$, low $C_L$ fluctuations, pressure drop and velocity magnitude, and a large recirculation area. In this way, efficient active control can reduce vortex shedding caused by the bluff cylinder and gain a smoother flow. Here, the reward function is defined as follows:
\begin{equation}
    r_t = -C_D - 0.2 |C_L|,
\end{equation}
which is to minimize the drag and lift to stabilize the vortex alley and the same as other papers to facilitate comparison. 

In the 2D simulation of the target system, our method is applied to learn the active flow control strategy and obtain attention weights of all 151 probes sensors simultaneously. We use the attention weight from the AE module when the D-AFS/PPO controller achieves stable rewards. Then, the probes with Top K weights are used for control with the same reward function and the original PPO controller. The $C_D$ and $C_L$ achieved with different methods are used for evaluations.

\subsection{Comparison with Other FS Methods}
\begin{figure*}
    % \centering
      \begin{minipage}[b]{1.0\textwidth}
       \includegraphics[width=1.0\textwidth]{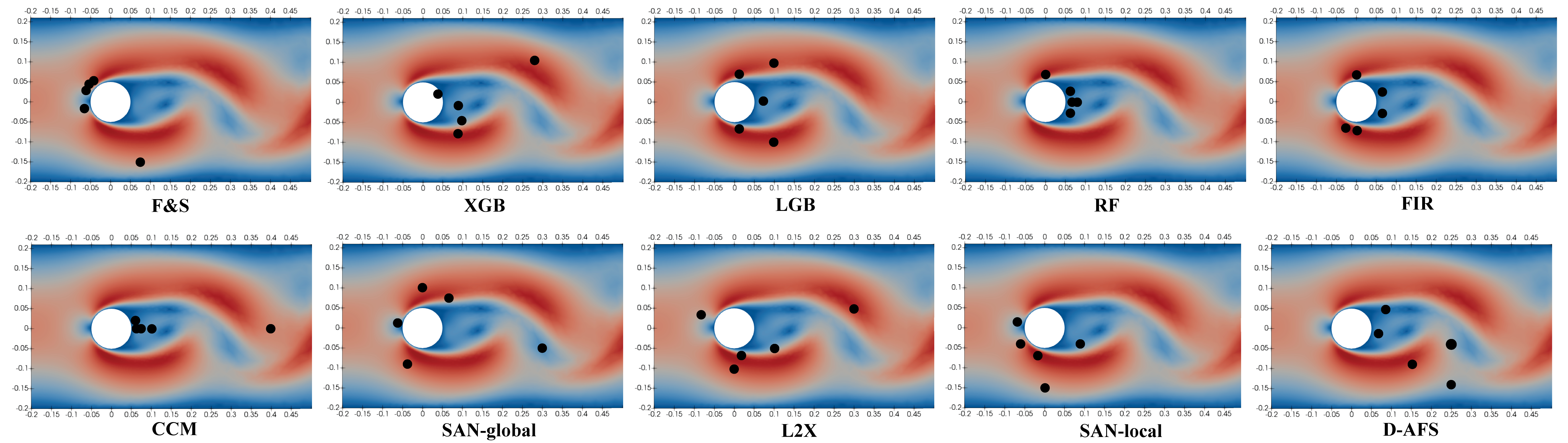}
       \end{minipage}%
 \vspace{0.3cm}      
\begin{minipage}[b]{0.9\textwidth}
\centering
    \begin{tabular}{cccccccccccc}
    \hline
        Methods & Uncontrolled & F\&S & XGB & LGB & RF & FIR & CCM & SAN-global & L2X & SAN-local & D-AFS\\
        \hline
        $C_D$ & 3.205 & 3.118 & 3.062 & 3.044 & 3.082 & 3.038 & 3.116 & 3.012 & 3.068 & 3.076 & \textbf{2.938} \\
        $C_L$ & 0.676 & 0.784 & 0.452 & 0.291 & 1.036 & 0.338 & 0.655 & 1.099 & 0.441 & 0.479  & \textbf{0.126} \\
        \hline
    \end{tabular}
% \end{table}
\end{minipage}%
\caption{Comparison of the Top 5 probes selected with different feature selection solutions. Table shows the average values of the drag coefficient $C_D$  and lift coefficient $C_L$  when the control strategy converges with only the selected five probes with the PPO controller}
\label{fig:fs_comparsions}
\end{figure*}

In this section, we test the performance of state-of-the-art baseline feature selection algorithms in their feature identification capabilities in control. We choose eight strong feature selection baselines, including classic methods, gradient boosting methods and recent population-wise methods. We also provide one instance-wise feature selection method. Implementation of these algorithms are either by the scikit-learn package\cite{pedregosa2011scikit} or from codes with their original papers. \textbf{F\&S} \cite{gu2011generalized}, short for Fisher Score, a similarity-based method, which selects each feature independently according to their scores under the fisher criterion. \textbf{XGB}\cite{chen2016xgboost}, short for XGBoost,  an optimized distributed gradient boosting library that can calculate the feature importance across many trees. \textbf{LGB}\cite{ke2017lightgbm}, short for LightGBM, gradient boosting framework that uses tree based learning algorithms. \textbf{RF}\cite{ho1995random}, short for Random Forest, operates by constructing a multitude of decision trees at training time. \textbf{FIR}\cite{wojtas2020feature}, a novel dual-net architecture consisting of operator and selector for discovery of an optimal feature subset. \textbf{CCM}\cite{chen2017kernel}, employs kernel-based measures of independence to find a subset of covariates that is maximally predictive of the response. \textbf{SAN}\cite{vskrlj2020feature}, explores the use of attention-based neural networks mechanism for estimating feature importance. We use both the local/global variants: SAN-local and SAN-global. \textbf{L2X}\cite{chen2018learning}, an instance-wise feature selection method that extract a subset of features that are most informative for each given example.

% \vspace{0.1in}
Time series sensory data and immediate rewards during the training and exploration of the PPO controller are collected to provide training data set for supervised feature selection. The data set includes 50 epochs series, with 80 control steps in each epoch.

Fig. \ref{fig:fs_comparsions} shows the Top 5 probes selected by different feature selection methods and corresponding control effects. The layouts selected by these supervised feature selection methods are not suitable for control, whether from numerical or physical perspectives. Most of them select probes whose characteristics do not change significantly during the periodic vortex shedding. Some of them, e.g. F\&S, SAN, and L2X,select the probes located on the left side of the cylinder, generally seen as an ineffective area. The value of these probes generally provides little help for the drag reduction task. In order to verify their effectiveness in control, we use PPO to control systems with only Top 5 selected probe, respectively.  The drag and lift coefficient in the table (bottom part of Fig. \ref{fig:fs_comparsions}) do clarify that these probes layouts cannot achieve effective active flow control behind the cylinder. In contrast, D-AFS achieve the best control effect for both $C_D$ and $C_L$ with the same number of probes. 

% \subsection{Results \& Analysis}
\subsection{Comparison with Existing DRL solutions}

% \noindent\textbf{Comparision with existing DRL solutions}

\begin{figure}[t]
    \centering
      \begin{minipage}[b]{0.48\textwidth}
       \includegraphics[width=0.99\textwidth]{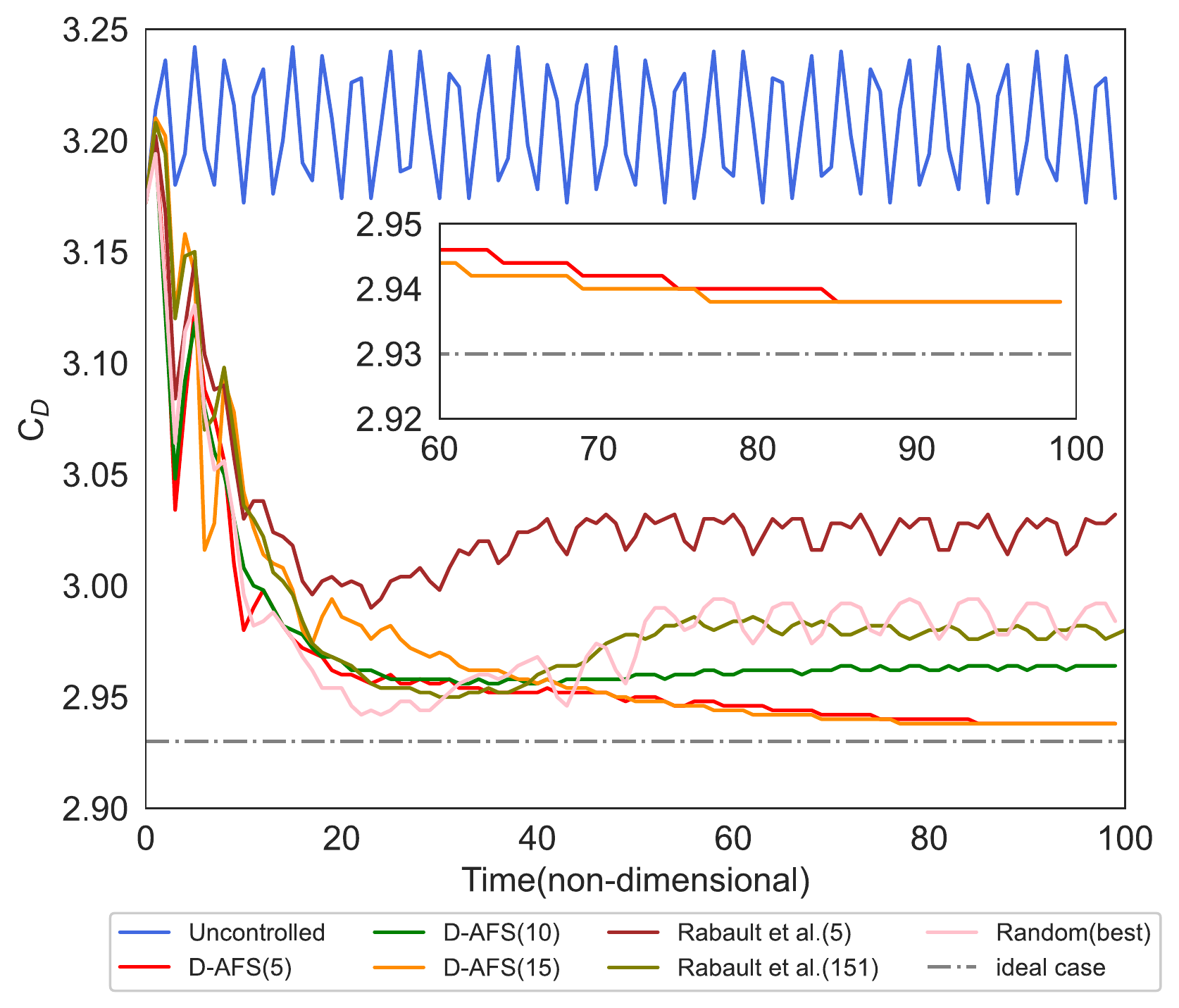}
    %   \caption{Image}
       \end{minipage}%
       
\begin{minipage}[b]{0.48\textwidth}
    \centering
    \begin{tabular}{ccccc}
    % \hline
    \toprule
        Methods & Probes & $C_D$ & $C_L$ & \makecell[c]{Rec Area} \\
        % \makecell[c]{Rec \\ Area}\\
        \midrule
        % \hline
        Uncontrolled & 151 & 3.205 & 0.676 & 0.012 \\
        Rabault et al.\shortcite{rabault2019artificial} & 151 & 2.980 & 0.210 & 0.022 \\
        % Tang et al. .\shortcite{tang2020robust} & 151 & 3.026 & 0.259 & 0.017 \\
        Rabault et al.\shortcite{rabault2019artificial} & 5 & 3.026 & 0.259 & 0.017 \\
        Random Selection & 5 & 2.987 & 0.424 & 0.019 \\
        \hline
        D-AFS(5) & 5 & \textbf{2.938} & 0.126 & \textbf{0.030} \\
        D-AFS(10) & 10 & 2.964 & 0.129 & 0.025 \\
        D-AFS(15) & 15 & \textbf{2.938} & \textbf{0.112} & 0.029 \\
        Ideal Case & - & 2.93 & - &  - \\
    % \hline
    \bottomrule
    \end{tabular}
   
    % \label{tab:sensor_placements}
% \end{table}
\end{minipage}%
% \vspace{-0.1in}
 \caption{Time-evolutionary value of the drag coefficient $C_D$  with different solution. Table shows the mean values of $C_D$, $C_L$ and Rec Area when system becomes steady.}
%  \vspace{-0.2in}
 \label{tab:sensor_placements}
\end{figure}

We verify the control performance obtained using the probe configurations selected by different DRL solutions with 151 probes and observe their results.
The performance of the active flow control strategy under different configurations is shown in the upper part of Fig.~\ref{tab:sensor_placements}. The numerical results for drag coefficient $C_D$, lift coefficient $C_L$ and recirculation area size are shown in the bottom part of Fig.~\ref{tab:sensor_placements}. In addition to TopK probe configurations, we also show the performance of \citeauthor{rabault2019artificial}'s result with selected five-probe, marked as Expert(5) and a randomly selected five-probe configuration denoted as Random(5). In Random(5), the best performance in five rounds is reported. Here, other reference use similar environment but different settings, e.g. Tang \citep{tang2020robust} use four actuators instead of two, thus the result can not be directly compared. Reference \cite{paris2021robust} requires heuristic enumeration of sensor layouts with only 15 probes and has not yet been open-sourced.

The top part of the figure shows that the drag coefficient $C_D$ varies periodically as the vortex shedding when the system is uncontrolled. The mean value of  $C_D$ is around 3.205, and the fluctuation amplitude is around 0.034. In comparison, taking the Top5/10/15 probes selected by D-AFS as observations, the PPO controller can achieve better active flow control. 
The $C_D$ of D-AFS with the five-sensor is very close to, more exactly 99.73\%, the ideal 2.93(denoted by dashed gray line) where no vortex shedding exists\cite{rabault2019artificial}. 
Compared with the five-sensor configuration by human experts\cite{rabault2019artificial}, D-AFS achieves a 260\% increase in the recirculation area and 49.2\% drag reduction. Even compared to the state-of-the-art solution with 151 probes, our five-sensor configuration can achieve a further 18.7\% drag reduction. 
The drag coefficient $C_D$ of the best random selection is better than Expert(5), but the lift coefficient $C_L$ is worse than the latter. That means the control strategy using randomly selected probes is unstable.

% In comparison, the human arbitrary selected by suffer the worst performance in all three 5 configurations, even much worse than Random(5). 

%  The strategies in 5 and 15 probes cases are relatively better.

% Furthermore, our results are very stable with little fluctuation. This fact shows that our solution can effectively weak the vortex.
Interestingly, adding more sensors to the five-sensor layout selected by D-AFS does not bring further drag reduction, which makes this five-sensor configuration the optimized trade-off between DRL performance and sensor selection complexity. It shows that too many possibly redundant inputs may cause interference and hindrance to the controller with limited modeling capabilities, albeit they might contain more information. This finding matches precisely the finding stated in \cite{paris2021robust}.

% some similar good control laws compared to the uncontrolled case. The drag is obviously reduced with the active flow control, almost the same as the full-featured training strategy.

\subsection{Further Analysis}
  In this section, we provide further analysis on the probe weight distribution, control effect analysis,  impacts of different probe configurations and the complexity analysis.
  
\begin{figure}[tb]
    \centering
    \includegraphics[width=0.48\textwidth]{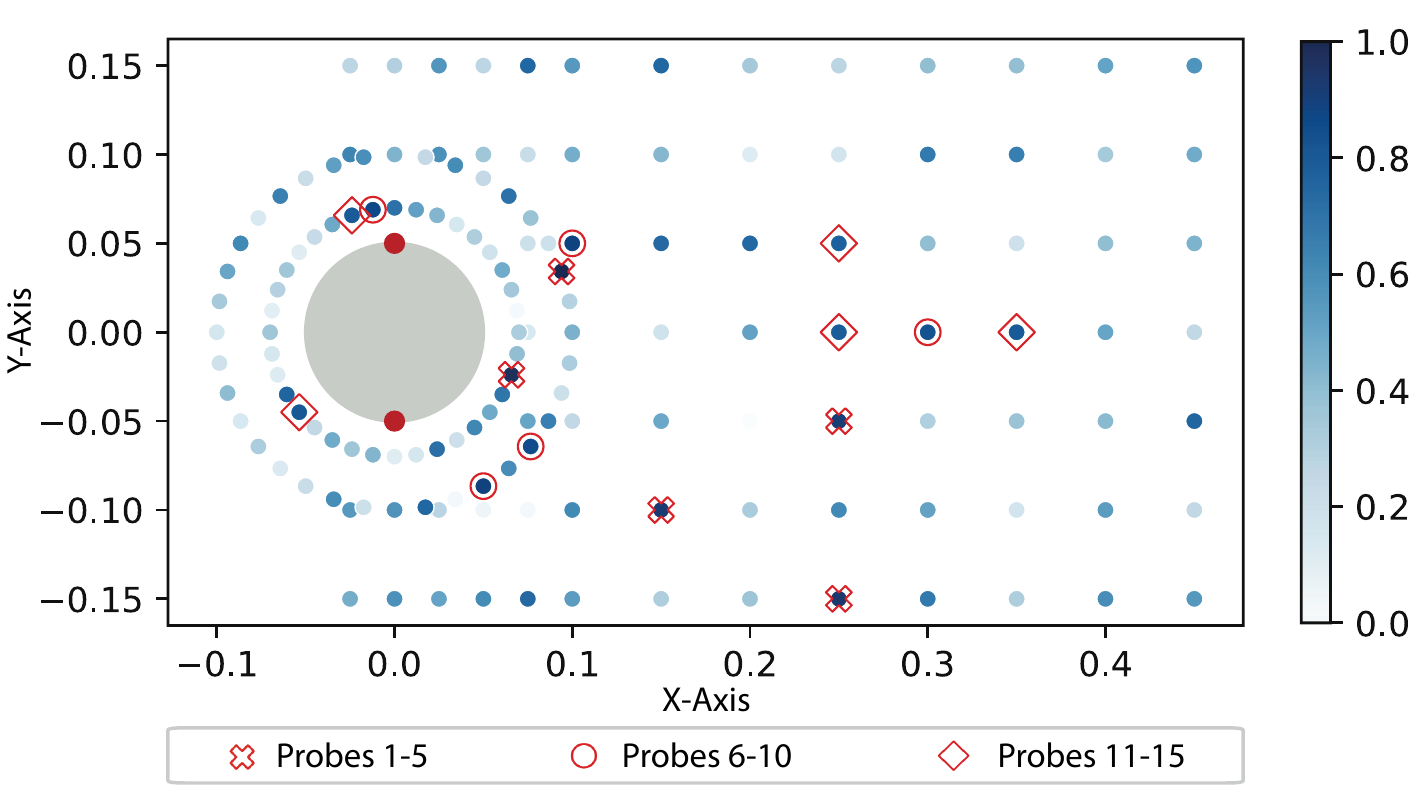}
    \caption{The attention weight distribution of 151 probes. The Top5/10/15 probes in weight ranking have been marked.}
    \label{fig:selection}
\end{figure}

\begin{figure}
\includegraphics[width=0.47\textwidth]{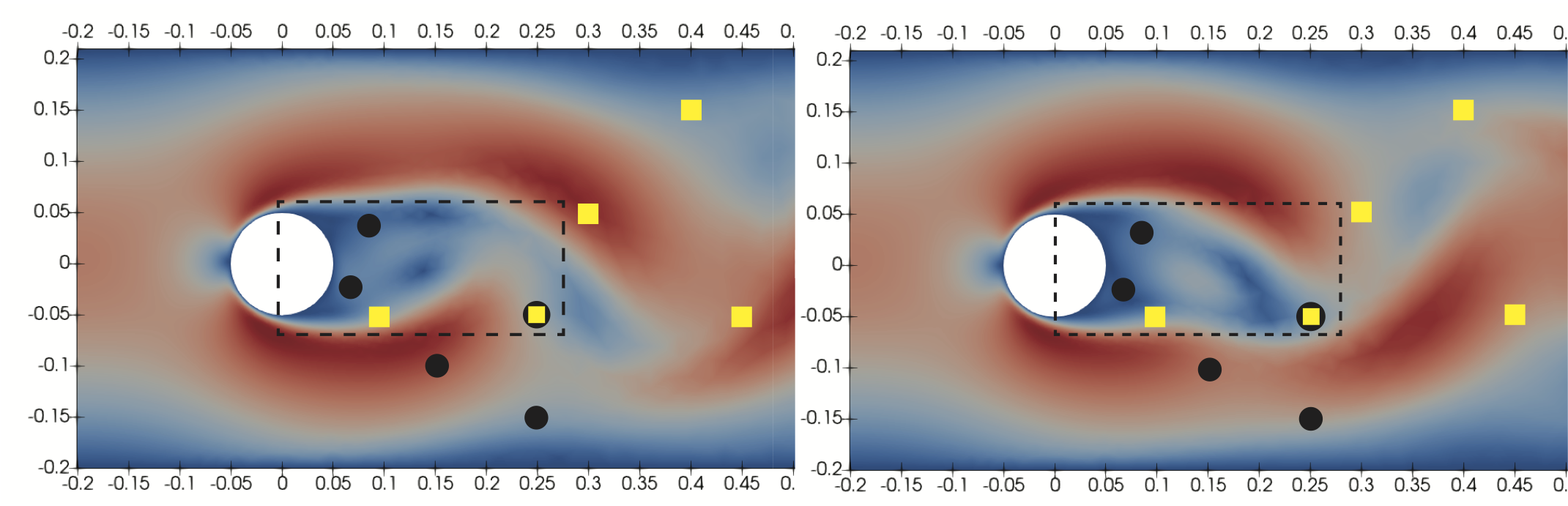}
% \vspace{-0.1in}
 \caption{Identified two equivalent probe layouts with similar DRL performance, black circles and yellow squares indicate the probes selected in two separated experiments.}
 \label{fig:equ_sensors}
 \vspace{-0.1in}
\end{figure}

\noindent\textbf{Feature weight distribution:} Fig.~\ref{fig:selection} illustrates the attention weight of 151 probes with different gradient colors. The darker the color, the bigger the weight and the higher the importance of the probe to the controller. We also marked the Top5, Top10, and Top15 probes with the cross, circle, and diamond. The Top5 probes selected by D-AFS are scattered behind the cylinder, and they provide an excellent `coverage' of the instantaneous vortex variation, as shown in Fig.~\ref{fig:equ_sensors}. These probes are normally critical for the control task, which can explain the selection preference of D-AFS to a certain extent. That is, probes that can deliver critical information conducive to learning will get a higher weight.
% This weight refers to the test result of the model saved under the training convergence.
% We have repeated experiments many times to achieve stable results. Although the probes selected by each experiment are different, the control strategies are very similar.

\begin{figure}[t]
    \centering
    \includegraphics[width=0.47\textwidth]{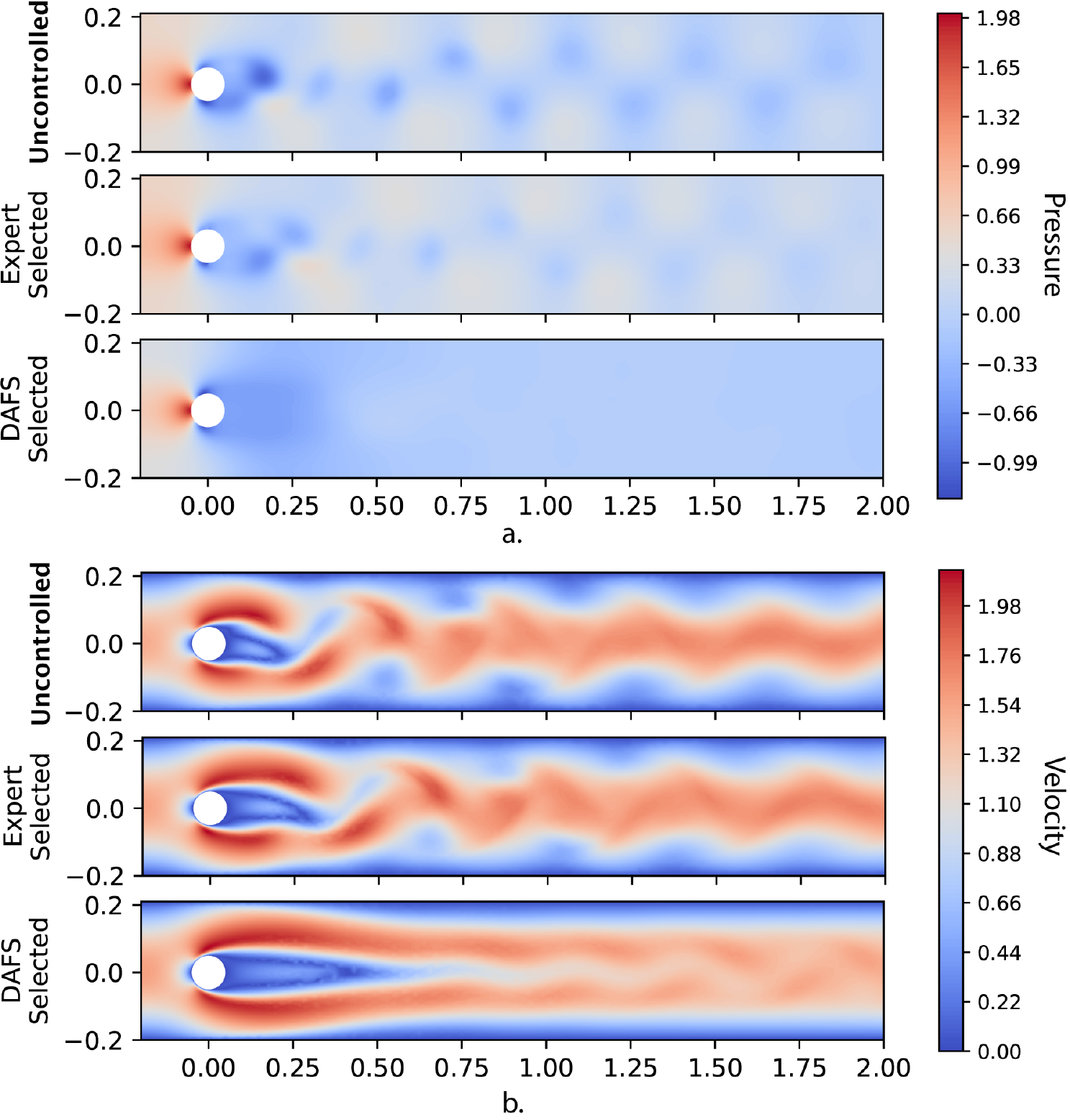}
    \caption{Comparison of active control performance(velocity and pressure) of uncontrolled and two five-probe configurations.}
    \label{fig:cfd_p_v}
    % \vspace{-0.1in}
\end{figure}

\noindent\textbf{Control effect analysis:} We take a look at macroscopic flow characteristics and how the active control affects them. Fig.~\ref{fig:cfd_p_v} shows the snapshots of the pressure (subgraph a.) and velocity distribution(subgraph b.) under three configurations when the system enters steady states. The three configurations are Uncontrolled, Expert(5), and D-AFS(5). 

% Based on the results of the previous subsection, it is found that D-AFS can automatically generate the quantitative importance of different observations through the exploration and exploitation process of DRL in the dual world.
% In this section, we will further pay attention to the optimal probe placement selection problem and expect to use partial system information provided by selected probes to get a better strategy. This experiment aims to inspect the control strategy using the partial observation selected by our method, which is mainly compared with the selection guided by human experts and random selection.

% We reproduce the paper's \cite{rabault2019artificial} result, both the complete observation case(151 probes, a relatively detailed flow description) and the partial observation case(5/11 probes manually selected by experts, which is marked in Figure \ref{fig:cfd_env}). 

The pressure distribution shows that the vortex in the wake caused by the cylinder is manifested as a significant pressure drop without active control. Similarly, the velocity magnitude fluctuates a lot, visible from the top sub-figure of the velocity distribution. The control effect from Expert(5) is shown in the middle of the two subgraphs, and we can see those probes are not in the key positions for vortex shedding. In comparison, the selected Top5 probes have good ``coverage'' of the instantaneous recirculation bubble. Thus, the control effect of D-AFS is much better. PPO with D-AFS(5) attenuates velocity fluctuations caused by the vortexes, achieves that vortexes are globally very weak and least active close to the upper and lower walls. More strikingly, the extent of the recirculation area is dramatically increased. 

\noindent\textbf{Equivalent probe layouts:}
As introduced previously, optimizing sensors' number and location are widely demanded in various industrial domains to find cost-effective sensor solutions in redundant probes. For D-AFS, it is possible to select probes with totally different layouts because of the random exploration and exploitation nature. Fig. \ref{fig:equ_sensors} shows two identified Top5 probes with two separated experiments, illustrated with black circles and yellow squares. They provide almost identical control performance. The drag coefficient $C_D$ is around 2.940, the lift coefficient $C_L$ is around 0.130.

The left and the right part of Fig. \ref{fig:equ_sensors} are two different snapshots, which show the forward propagation of the vortex to better illustrate selected probes' positions on the route of vortex. We can find that, although the two configurations are highly different(only one overlap), each configuration covers the key positions which contain richer information of periodic vortex shedding or located around the edge of the recirculation area. For instance, observations detected by probes located at (0.15, -0.1), (0.25, -0.15), (0.3, 0.05), (0.4,0.15) and (0.45, -0.05) enable the DRL controller obtain rich periodic information. This also means that the control task can be converted from drag reduction to vibration reduction for these observations.

\noindent\textbf{Complexity analysis:} Table~\ref{tab:complexity} shows the time complexity in generating feature weights for systems with different numbers of probes, ranging from 5 to 151. We also show the time when pure PPO is used for training. The settings for epochs and steps are the same as the experiment settings mentioned earlier. From this table, we can see that D-AFS generates feature weights in the process of DRL exploration and exploitation, and the overall time consumed is almost the same as that of pure PPO. The computation and resource complexity is relatively low. 

In comparison, if the DRL is only used as an arbiter with a certain heuristic-guided probe configuration, e.g., \cite{paris2021robust}, the computation complexity is about $O(C_n^r)$ with n are possible probes, and $r$ are probes to select. It will increase rather fast when n and r increase. The author pointed out that each training costs around 40 CPU hours when the system only has 12 probes. Although they provide no discussion on computation complexity, we deduce that this approach can hardly be applied to the system with 151 probes. 

\begin{table}[ht]
\caption{The computational complexity of D-AFS and pure PPO without feature selection.(Time unit: hour)}
    \centering
    
    \setlength{\belowdisplayskip}{0.5cm}
    \begin{tabular}{c|ccccc}
    \hline
        Probes & 5 & 10 & 50 & 100 & 151  \\
    \hline
        PPO & 3.17 & 3.24 & 3.30 & 3.39 & 3.46 \\
        % \hline
        D-AFS/PPO & 3.19 & 3.28 & 3.37 & 3.49 & 3.60 \\
        \hline
    \end{tabular}
    \label{tab:complexity}
    % \vspace{-0.2in}
\end{table}
% \vspace{-0.2cm}

\subsection{Further Experiments in OpenAI Gym}

To better explore the generalization of D-AFS, we also tune two popular DRL algorithms, DQN and DDPG into our framework. We choose four classical control cases provided by OpenAI Gym\cite{brockman2016openai}, including Pendulum, MountainCar, CartPole, and Acrobot, to observe control performance with feature subsets selected by D-AFS.

% We also tune two popular DRL algorithms, DQN and DDPG\cite{lillicrap2016continuous} into our D-AFS framework. A set of experiments has been performed on four classical control cases provided by OpenAI Gym\cite{brockman2016openai}, including Pendulum, MountainCar, CartPole, and Acrobot. We use the original inputs and certain transformation functions to generate a set of new constructed features and one random noise as possible inputs. We set the same network and training configuration for DQN as the ones defined in Gym. The data constitutes 200000 steps of interaction data between the controller and the system which cover all the interaction process information between the controller and the system.

\begin{figure}[hbp]
    \centering
    \includegraphics[width=0.47\textwidth]{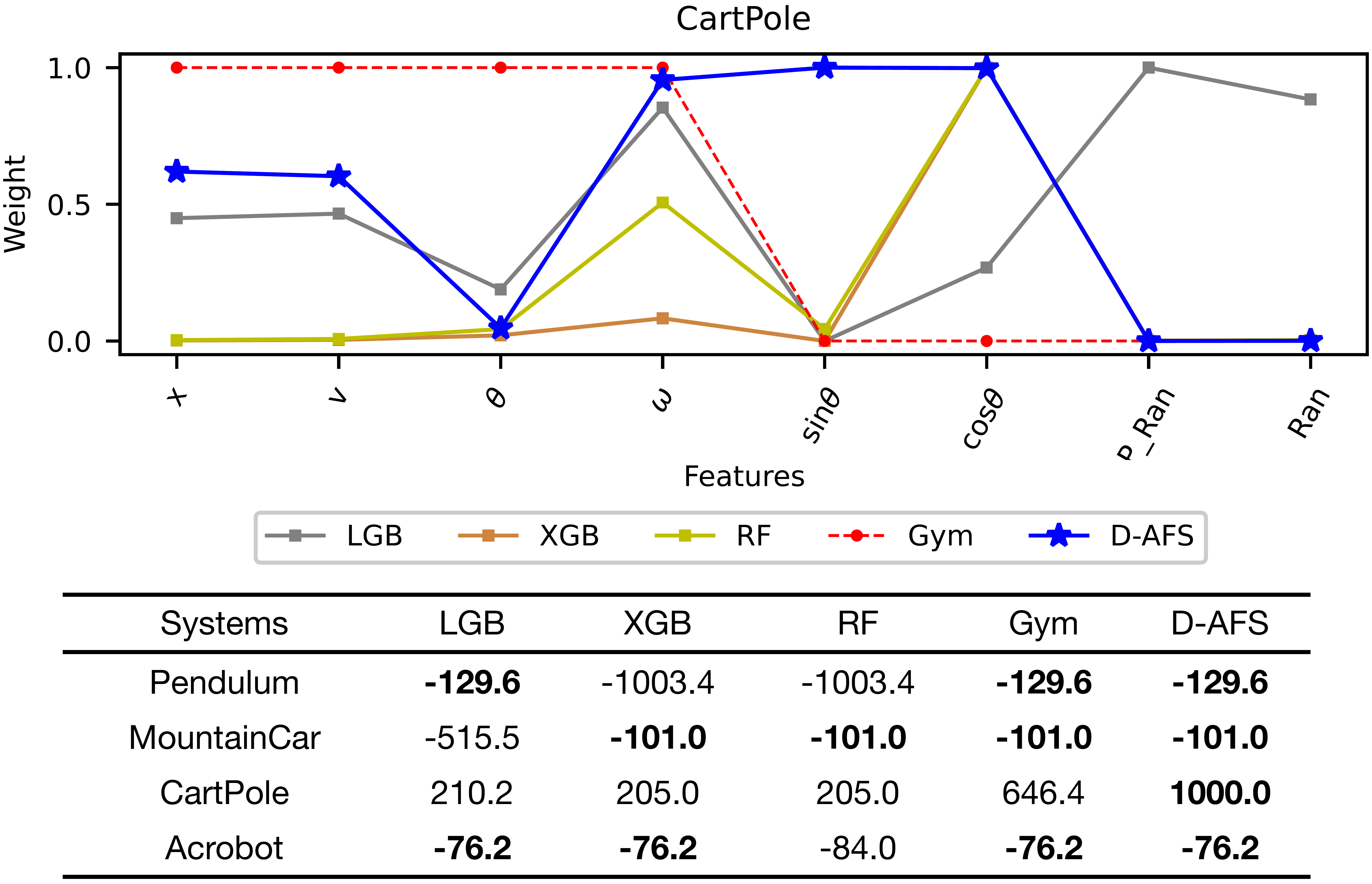}
    \caption{Comparison of controller performance with the feature subsets selected by different methods.}
        % \vspace{-0.2in}
    \label{fig:weight_diff}
\end{figure}

The lower part of Fig.~\ref{fig:weight_diff} is a table, in which values are rewards of four systems with feature subsets selected by different methods. The larger the rewards, the better the performance of DRL's controller. 
To verify that the control performance is influenced only by features, experimental settings are all the same except for the feature subsets.
Results in the table show that D-AFS can select relevant features effectively and achieve the same or even better control performance than the original expert-based subsets. In Pendulum, MountainCar and Acrobot, D-AFS achieves the same rewards as Gym. Actually, it deduces the same sensor sets used with Gym in all three cases. In Cartpole, D-AFS achieves the maximum return under the limit of 1000 maximal steps in an episode, which is about 54.70\% higher than the 646.4 achieved by the features provided by Gym with a different set of features. Due to the page limits, the upper part of Fig.~\ref{fig:weight_diff} only shows the feature weights calculated by different methods for CartPole. Since the feature weight values generated by different methods are not uniform, we normalize all weights for better comparison. Regarding the features provided by Gym based on human expertise, we set their weights to 1, while other features that are not included are set to 0. More comprehensive weights for other cases are provided in Appendix A.

%  show the feature weights generated by different methods. Since the feature weight values generated by different methods are not uniform, we normalize all weights for better comparison. Regarding the features provided by Gym based on human expertise, we set their weights to 1, while other features that are not included are set to 0. The table in the lower part shows the maximal rewards achieved with the set of features selected via different methods.

% Results in the table further verify our findings of the limitations of traditional FS for control. Even for those four comparable simple systems, XGB and LGB can not effectively find most relevant features. In addition, their performance in different systems is also volatile. 
% % XGB and RF can effectively eliminate noise signals, but they cannot find the necessary features for control. Moreover, the feature weights generated by these ML methods sometimes are very indistinguishable from each other, such as the performance of XGB in CartPole. 
% The poor cumulative return in the table can also illustrate their limitations in control.
% These facts show that D-AFS can effectively select the most relevant features for control, which is crucial for understanding the environments without a model. Due to page limits, comprehensive experiment results are provided in Appendix A. 

\section{Conclusion}
Unlike humans who are good at learning unknown environments through interaction, the most existing DRL algorithms cannot gain explicit environmental knowledge from the learning and exploration process. In this paper, a dual-world feature selection method D-AFS is proposed to quantitatively evaluate the impacts of different inputs for control during the DRL training process. A set of experiments is performed in one classic flow control. Results show that D-AFS can effectively identify the system-mechanism-related inputs. With the deduced five-probe layout, we can achieve the best drag reduction than all other strong baselines. Results highlight, we argue, a new way to achieve efficient and optimal sensor designs for experimental or industrial systems. In this paper, our work limits to the active flow problem with limited complexity due to the high computation complexity with FEniCS for solving partial differential equations. We are working on implementing D-AFS on a deep neural network based simulator for more fast and complex tasks. The other possible direction is to find a solution that can automatically decide the optimal set of sensors selected for control. 
% \textcolor{blue}{add limitation about cannot decide the optimal number of sensors.}

% The proposed method provides an automatic and feasible way to understand unknown systems with the DRL's learning capabilities.

% There remain some questions to be answered in future work. Currently, we only focus on the average weight distributions of different features. It is possible, in the specific scenarios, different inputs might have scenario-specific weights. Those temporal effects sometimes have profound means for the system dynamics interpretations. Another direction for the D-AFS is the better integration of attention-weighted mechanisms with more DRL algorithms.

\section{Acknowledgement}
This work was supported by the National Science Foundation of China (61772473, 62073345 and 62011530148). F. Wang was supported by the Equipment Pre-research Key Laboratory Funds No. 61425010102.

% \begin{verbatim}
  \bibliographystyle{ACM-Reference-Format}
  \bibliography{kdd22}
% \end{verbatim}

\onecolumn
\section*{Appendix A:  The experimental results of OpenAI Gym}
Since our method has good scalability and can minimize the changes to the existing DRL algorithm, we apply the architecture to two mainstream algorithms, DQN and DDPG.
We apply D-AFS/DQN and D-AFS/DDPG to four classical control environments provided by OpenAI Gym, the Pendulum, MountainCar, CartPole, and Acrobot.
The four target systems are shown in Fig.~\ref{fig:system}, and the weights generated by D-AFS are shown in Table \ref{D-AFS_selection}. The boldface in the table indicates the features selected by D-AFS. ``P\_Ran" and ``Ran" represent partial noise and complete noise, respectively. The results show that our method can effectively select the most relevant sensors and reduce human expertise reliance.

\label{sec:appendixA}
% \FloatBarrier

% \begin{table*}[hbp]
% \renewcommand{\arraystretch}{1.3}
% \caption{The four classical control systems provided by OpenAI Gym}
% \label{table_system}
% \resizebox{\textwidth}{!}{
% \centering
% \begin{tabular}{ccccc}
% \hline
% \bfseries Systems & \bfseries Raw inputs & \bfseries Gym inputs & \bfseries Constructed inputs & \bfseries Reward functions\\
% \hline
% Pendulum & $\theta$ & $\cos\theta$, $\sin\theta$, $\omega$ & $\theta$, $\cos\theta$, $\sin\theta$, $\omega$, P\_Ran, Ran & reward$=\theta^{2} + 0.1\omega^{2} + 0.001u^{2}$\\
% MountainCar & $x$, $v$ & $x$, $v$ & $x$, $v$, P\_Ran, Ran & \makecell[c]{Discrete: if not $x>0.5$, reward=$-1.0$\\Continuous: reward$-=0.1u^2$,if $x>0.5$, reward=100.0}\\
% CartPole & $x$, $\theta$ & $x$, $v$, $\theta$, $\omega$ & $x$, $v$, $\theta$, $\omega$, $\sin\theta$, $\cos\theta$, P\_Ran, Ran & $\forall(abs(x)<2.4\land abs(\theta)<12^{\circ})$, reward=$1.0-x^2$\\
% % \tabincell{c}{Discrete:$\forall(abs(x)<2.4\land abs(\theta)<12^{\circ})$, reward+=$1.0$ \\ Continuous:$\forall(abs(x)<2.4\land abs(\theta)<12^{\circ})$, reward=$1.0-x^2$} \\
% Acrobot & $\theta_1$, $\theta_2$ & \makecell[c]{$\sin\theta_1$, $\cos\theta_1$, \\ $\sin\theta_2$, $\cos\theta_2$, \\ $\omega_1$, $\omega_2$} & \makecell[c]{$\theta_1$, $\theta_2$, $\sin\theta_1$, $\cos\theta_1$, $\sin\theta_2$, \\$\cos\theta_2$, $\omega_1$, $\omega_2$, P\_Ran, Ran} & $\forall(-\cos\theta_1-\cos(\theta_1+\theta_2))<=1.0$, reward=$-1.0$\\
% \hline
% \end{tabular}}
% \end{table*}
\setcounter{figure}{0}
\renewcommand{\thefigure}{A\arabic{figure}}
\begin{figure*}[hbp]
    \centering
    \includegraphics[width=0.95\textwidth]{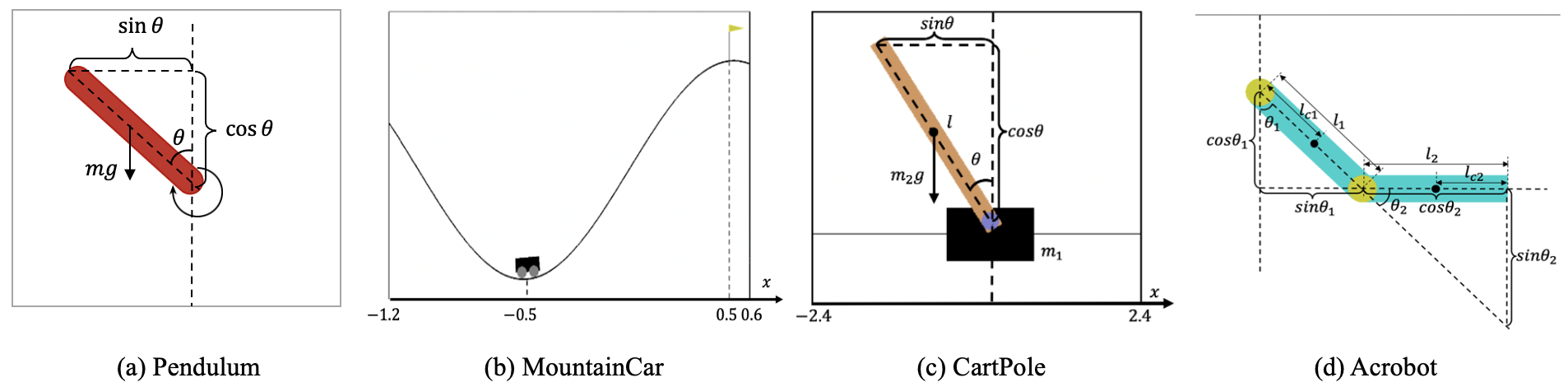}
    \caption{The physical meaning of the major features in the four systems}
    \label{fig:system}
\end{figure*}

\setcounter{table}{0}
\renewcommand{\thetable}{A\arabic{table}}
% \FloatBarrier
\begin{table*}[hb]
\renewcommand{\arraystretch}{1.3}
\caption{Results of D-AFS in four classical control systems provided OpenAI Gym}
\label{D-AFS_selection}
\resizebox{\textwidth}{4cm}{
\centering
\begin{tabular}{cccccccccccc}
\hline
\bfseries Algorithm & \bfseries Systems & \multicolumn{9}{c}{\textbf{All inputs and weights}} \\
\hline
\multirow{9}*{D-AFS/DQN} & \multirow{2}*{Pendulum} & $\theta$ & $\omega$ & $\sin\theta$ & $\cos\theta$ & P\_Ran & Ran & ~ & ~ & ~ & ~ \\
		~ & ~ & 0.516 & \textbf{0.968} & \textbf{0.960} & \textbf{0.879} & 1.27e-4 & 2.32e-5& ~ & ~ & ~ & ~ \\
~ & \multirow{2}*{MountainCar} & $x$ & $v$ & P\_Ran & Ran & ~ & ~ & ~ & ~ & ~ & ~ \\
~ & ~ & \textbf{0.862} & \textbf{0.828} & 7.99e-5 & 1.75e-4 & ~ & ~ & ~ & ~ & ~ & ~ \\
~ & \multirow{2}*{CartPole} & $x$ & $v$ & $\theta$ & $\omega$ & $\sin\theta$ & $\cos\theta$ & P\_Ran & Ran & ~ & ~ \\
~ & ~ & \textbf{0.618} & \textbf{0.601} & 0.047 & \textbf{0.952} & \textbf{0.997} & \textbf{0.955} & 2.28e-4 & 9.57e-4 & ~ & ~ \\
~ & \multirow{3}*{Acrobot} & $\theta_1$ & $\theta_2$ & $\omega_1$ & $\omega_2$ & $\sin\theta_1$ & $\cos\theta_1$ & $\sin\theta_2$ & $\cos\theta_2$ & P\_Ran & Ran \\
~ & ~ & \textbf{0.907} & 0.407 & \textbf{0.921} & \textbf{0.470} & 0.356 & \textbf{0.965} & \textbf{0.809} & \textbf{0.848} & 1.04e-3 & 1.95e-3 \\
~ & ~ & 0.408 & 0.427 & \textbf{0.934} & \textbf{0.537} & \textbf{0.674} & \textbf{0.938} & \textbf{0.658} & \textbf{0.924} & 1.75e-3 & 2.79e-3 \\
\hline
\multirow{8}*{D-AFS/DDPG} & \multirow{2}*{Pendulum} & $\theta$ & $\omega$ & $\sin\theta$ & $\cos\theta$ & P\_Ran & Ran & ~ & ~ & ~ & ~ \\
		~ & ~ & 0.243 & \textbf{0.561} & \textbf{0.956} & \textbf{0.739} & 1.20e-3 & 2.96e-3& ~ & ~ & ~ & ~ \\
~ & \multirow{2}*{MountainCar} & $x$ & $v$ & P\_Ran & Ran & ~ & ~ & ~ & ~ & ~ & ~ \\
~ & ~ & \textbf{0.178} & \textbf{0.894} & 3.48e-4 & 3.27e-4 & ~ & ~ & ~ & ~ & ~ & ~ \\
~ & \multirow{2}*{CartPole} & $x$ & $v$ & $\theta$ & $\omega$ & $\sin\theta$ & $\cos\theta$ & P\_Ran & Ran & ~ & ~ \\
~ & ~ & \textbf{0.822} & 0.386 & \textbf{0.955} & 0.356 & \textbf{0.868} & \textbf{0.607} & 0.009 & 0.002 & ~ & ~ \\
~ & \multirow{2}*{Acrobot} & $\theta_1$ & $\theta_2$ & $\omega_1$ & $\omega_2$ & $\sin\theta_1$ & $\cos\theta_1$ & $\sin\theta_2$ & $\cos\theta_2$ & P\_Ran & Ran \\
~ & ~ & \textbf{0.651} & \textbf{0.459} & 0.273 & 0.302 & \textbf{0.609} & \textbf{0.754} & \textbf{0.785} & \textbf{0.821} & 2.67e-3 & 2.16e-3 \\
\hline
\end{tabular}}
\end{table*}

\end{document}